\documentclass{article}

     \PassOptionsToPackage{numbers, compress}{natbib}

    \usepackage[preprint]{neurips_2020}

\usepackage[utf8]{inputenc} 
\usepackage[T1]{fontenc}    
\usepackage{hyperref}       
\usepackage{url}            
\usepackage{booktabs}       
\usepackage{amsfonts}       
\usepackage{nicefrac}       
\usepackage{microtype}      

\usepackage[ruled]{algorithm2e}
\usepackage{graphicx, xcolor}
\usepackage{tikz}
\usepackage{caption}
\usepackage{subfig}
\usepackage{amsmath,amsgen,amstext,amssymb}
\usepackage{amsfonts, amsthm}
\usepackage{dsfont}
\usepackage{multicol}
\usepackage{mathtools}
\usepackage{esint}

\SetKwProg{Fn}{Function}{}{}

\DeclareMathAlphabet{\mathpzc}{OT1}{pzc}{m}{it}
\DeclareMathOperator*{\argmax}{\arg\max}

\newtheorem{theorem}{Theorem}[section]

\newtheorem{claim}{Claim}[section]

\newcommand{\bB}{{\mathbf{B}}}
\newcommand{\bC}{{\mathbf{C}}}
\newcommand{\bD}{{\mathbf{D}}}

\newcommand{\bI}{{\mathbf{I}}}

\newcommand{\bN}{{\mathbf{N}}}

\newcommand{\bT}{{\mathbf{T}}}

\newcommand{\bb}{{\mathbf{b}}}

\newcommand{\bu}{{\mathbf{u}}}

\newcommand{\bx}{{\mathbf{x}}}

\newcommand{\bbA}{{\mathbb{A}}}

\newcommand{\bbQ}{{\mathbb{Q}}}

\newcommand{\bbX}{{\mathbb{X}}}

\newcommand{\var}{\mathbb{V}\mbox{ar}}

\newcommand{\ex}{\mathbb{E}}
\newcommand{\pr}{\mathbb{P}}

\newcommand{\beqn}{\begin{equation}}
\newcommand{\eeqn}{\end{equation}}

\newcommand{\cF}{{\mathcal{F}}}

\newcommand\Reals{{\mathbb{R}}}
\newcommand\Ints{{\mathbb{Z}}}

\newcommand{\setA}{{\bbA}}
\newcommand{\setQ}{{\bbQ}}

\newcommand{\setX}{{\bbX}}
\newcommand{\rewardR}{{r}}
\newcommand{\rewardC}{{c}}
\newcommand{\BellmanOp}{{\bB}}

\newcommand{\NeighborOp}{{\bN}}
\newcommand{\ContractionOp}{{\bT}}

\title{A General Markov Decision Process Framework for Directly Learning Optimal Control Policies}

\author{%
  Yingdong Lu, Mark S.~Squillante, Chai W.~Wu \\
  Mathematical Sciences\\
  IBM Research\\
  Yorktown Heights, NY 20198, USA\\
  \texttt{\{yingdong, mss, cwwu\}@us.ibm.com} \\
}

\begin{document}

\maketitle

\begin{abstract}
We consider a new form of reinforcement learning (RL) that is based on opportunities to directly learn the optimal control policy
and a general Markov decision process (MDP) framework devised to support these opportunities. Derivations of general classes of our control-based RL methods are presented, together with forms of exploration and exploitation in learning and applying the optimal control policy over time.
Our general MDP framework extends the classical Bellman operator and optimality criteria by generalizing the definition and scope of a policy for any given state.
We establish the convergence and optimality~--~both in general and within various control paradigms (e.g., piecewise linear control policies)~--~of our control-based methods through this general MDP framework, including convergence of $Q$-learning within the context of our MDP framework.
Our empirical results demonstrate and quantify the significant benefits of our approach.
\end{abstract}

\section{Introduction}
Over the past many years, reinforcement learning (RL) has proven to be very successful in solving a wide variety of learning and
decision making
problems.
This includes problems
related to game playing
(e.g.,~\cite{Tesa95,ToKaKo+09}),
bicycle riding
(e.g.,~\cite{RanAls98}),
and robotic control
(e.g.,~\cite{RiGaHa+09}).
Many different RL approaches, with varying levels of success, have been developed to address these problems \cite{KaLiMo96,Szep10,SutBar11}.

Among these different approaches,
model-free RL has been demonstrated to 
solve various problems without any prior knowledge
(e.g.,~\cite{RanAls98,MnKaSi+13}).
Such model-free approaches, however, often suffer from high sample complexity that can require an inordinate amount of samples for some problems
which can be prohibitive in practice, especially for
problems limited by time or other constraints.
Model-based RL has been demonstrated to significantly reduce sample complexity and
shown to outperform model-free approaches
for various problems
(e.g.,~\cite{DeiRas11,MeHiXu+15}).
Such model-based approaches, however, can often suffer from the difficulty of learning an appropriate model and from worse asymptotic
performance than model-free approaches due to model bias from inherently assuming the learned system dynamics model accurately represents
the true system environment
(e.g.,~\cite{AtkSan97,Schn97,Scha97}).

In this paper we propose a novel form of RL that seeks to directly learn an optimal control model for a general underlying (unknown) dynamical system and to directly apply the corresponding optimal control from the model.
This is in strong contrast to many traditional model-based RL methods that, after learning the system dynamics model which is often of high complexity and dimensionality, then use this system dynamics model to compute an optimal solution of a corresponding dynamic programming problem, often applying model predictive control
(e.g.,~\cite{NaKaFe+17}).
Our control-based RL approach instead learns the parameters of an optimal control model, often of lower complexity and dimensionality, from which the optimal solution is directly obtained.
Furthermore, we establish that our control-based RL approach converges to an optimal solution analogous to model-free RL approaches while eliminating the problems of model bias in traditional model-based RL approaches.

We also introduce a general Markov decision process (MDP) framework that extends the classical Bellman operator and optimality criteria by generalizing the definition and scope of a policy for any given state and that provides theoretical support for our general control-based RL approach.
Within this framework, we establish results on convergence with respect to (w.r.t.) both a contraction operator and a corresponding form of $Q$-learning, and then establish results on various aspects of optimality and optimal policies.
%
%
To the best of our knowledge, this is the first proposal and analysis of a general control-based RL approach together with theoretical support from an extended MDP framework,
both of which
should be exploited to a much greater extent in the RL literature.

We first present our general control-based RL approach that directly learns the parameters of 
the optimal control models, including the corresponding convergence result.
Our general MDP framework is then presented together with the corresponding theoretical results on convergence and optimality.
Lastly, we present empirical results for a few classical problems in the OpenAI Gym framework \cite{AIgym} that demonstrate and quantify
the significant benefits of our general approach over existing algorithms.
We will release on GitHub the corresponding python code and make it publically available.


We refer to the Appendices in general for detailed proofs, additional technical details and results.

\section{Control-Based Reinforcement Learning}
\label{sec:controlRL}
Consider a discrete-state, discrete-time RL framework
defined over a set of states $\setX$, a set of actions $\setA$, a transition probability kernel $\pr$, a
cost function $\rewardC$ or a
reward function $\rewardR$
mapping state-action pairs to a bounded subset of $\Reals$, and a discount factor $\gamma \in [0,1)$.
Let $\ex_{\pr}$ denote expectation
w.r.t.\ 
the probability kernel $\pr$.
Then the discounted infinite-horizon stochastic dynamic programming (DP) formulation for the corresponding RL problems of interest can be expressed as
\begin{equation}
\min_{a_1, a_2, \ldots} \;\;  \;\; \ex_\pr\left[ \sum_{t=0}^\infty \gamma^t c(x_t, a_t, x_{t+1}) \right] , \label{eqn:general_control:min} 
\qquad \mbox{s.t.} \;\;  \;\; x_{t+1} = f(x_t, a_t), 
\end{equation}
where $x_t \in \setX$ represents the state of the system, $a_t \in \setA$ represents the control action decision variable, $f(\cdot, \cdot)$ represents the evolution function of the dynamical system characterizing the system state given the previous state and the taken action together with unknown uncertainty,
and $c(\cdot, \cdot, \cdot)$ represents a cost-based objective function of both the system states and control action.
Alternatively, the stochastic DP formulation associated with the RL problem of interest can be expressed as
\begin{equation}
\max_{a_1, a_2, \ldots} \;\;  \;\; \ex_\pr\left[ \sum_{t=0}^\infty \gamma^t r(x_t, a_t, x_{t+1}) \right] , \label{eqn:general_control:max} 
\qquad \mbox{s.t.} \;\;  \;\; x_{t+1} = f(x_t, a_t), 
\end{equation}
where $r(\cdot, \cdot,\cdot)$ represents a reward-based objective function of both the system states and control action, with all other variables
and functions as described above.

We note that \eqref{eqn:general_control:min} and \eqref{eqn:general_control:max}
can represent a wide variety of RL problems and corresponding stochastic DP problems based on the different forms taken by the evolution function $f(\cdot,\cdot)$, together with the transition probability kernel $\pr$.
For example, a linear system dynamics model results when $f(\cdot,\cdot)$ takes the form of linear transformations.
The function $f(\cdot,\cdot)$ can also characterize the discretized evolutionary system dynamics governed by (stochastic) differential equations or
(stochastic) partial differential equations.
In addition, the cost function $c(\cdot,\cdot,\cdot)$ and reward function $r(\cdot,\cdot,\cdot)$ are also allowed to take on various general forms, and thus
can represent any combination of cumulative and terminal costs or rewards, respectively.
Both formulations \eqref{eqn:general_control:min} and \eqref{eqn:general_control:max}
can also be analogously defined in continuous time.
On the other hand, most classical RL formulations assume the dynamics evolve in discrete time.
When the underlying system is based on a continuous-time model, then a discretization operator such as forward Euler discretization can be used to generate the discrete time samples.

Traditional model-based RL methods seek to first learn the system dynamics model of high complexity and dimensionality, and then incur the additional overhead of computing a solution to \eqref{eqn:general_control:min} or
\eqref{eqn:general_control:max} to obtain an approximation of the optimal control policy
w.r.t.\
the learned system dynamics model.
In strong contrast, we take an optimal control-theoretic perspective and seek opportunities to directly learn an  optimal control model of lower complexity and dimensionality for a general underlying dynamical system, and then directly apply the corresponding
optimal control policy from the model.

Consider a control policy with a linear or simple nonlinear form, respectively given by
\begin{equation}
a(y)  =  B y + b
\qquad \mbox{ and } \qquad
a(y)  =  B y + C y^T Dy + b ,
\label{eq:lin-quad-control}
\end{equation}
where $y = K(x)$ is the output process representing a measurement of the state variable (which can be linear or nonlinear), $B, C, D, b$ are matrices or vectors with the proper dimensions, and $T$ denotes the transpose operator.
We then need to learn $B, b$ in the case of the linear control model, and learn $B, C, D, b$ in the case of the simple nonlinear control model,
based on the RL sample measurements satisfying the corresponding expressions above.
Various
forms of RL can be applied to determine the best parameters for $B, C, D, b$.
In particular, we can exploit the low complexity and dimensionality of learning the parameters of the optimal control model, especially relative to the high complexity
and dimensionality of learning the system dynamics model, to solve the corresponding optimization problem after a relatively
small number of sample measurements.
Hence, the learning problem is reduced to solving a relatively small stochastic optimization problem in which uncertainty has to be sampled.
Many different algorithms
can be deployed to solve these
optimization problems and directly learn the desired control policy based on forms of exploration and exploitation.


More generally, our approach straightforwardly extends in an analogous manner to much more complicated nonlinear controllers, as well as piecewise linear and piecewise nonlinear controllers that can be considered as higher-order approximations of the optimal controller.
%

Our general control-based approach can be exploited together with many different forms of RL.
We next summarize a generic simplified version of our
approach.
%
\begin{enumerate}
\item Identify the reduced complexity and dimensionality control model based on the dimensions of the state and action vectors, possibly exploiting
additional prior knowledge.
\item Initialize  $B^*=B_0, b^* = b_0$ (linear) or $B^*=B_0, C^*=C_0, D^*=D_0, b^* = b_0$ (nonlinear).
\item Execute the system for the $e$-th episode of the RL task using the current control-model parameters $B_e, b_e$ (linear) or $B_e, C_e, D_e, b_e$ (nonlinear), thus
generating a sequence of state, action, reward tuples. 
\item When the episode yields is an improvement in the total reward, update the control-model parameters $B^*=B_e, b^* = b_e$ (linear) or $B^*=B_e, C^*=C_e, D^*=D_e, b^* = b_e$ (nonlinear).
\item Identify an alternative candidate for the control model $B_{e+1}, b_{e+1}$ (linear) or $B_{e+1}, C_{e+1}, D_{e+1}, b_{e+1}$ (nonlinear),
based on one of numerous available options.
\item Increment $e$ and repeat from Step $3$ until $B^*,b^*$ or $B^*, C^*, D^*,b^*$ satisfy some criteria.
\end{enumerate}

Prior knowledge (e.g., expert opinion, mathematical models), when available, can also be exploited in Step $1$ to bound the degrees of freedom of the problem;
in such cases, we simply take advantage of the additional prior knowledge and boundary conditions available in many real-world problems.
Initial conditions can be determined from prior solutions of similar problems, determined mathematically from a simple model, or chosen randomly.
Step $3$ consists of evaluating the total reward of each episode $e$.
An alternative candidate for the control model can be identified in numerous ways based on various forms of exploration and exploitation.
While the above generic simplified version of our control-based RL approach can exploit many different algorithms of interest, we focus in this paper on a general class of optimal search algorithms to efficiently and effectively learn the control models, for which we establish below asymptotic convergence.
%
We further note that, although our discussion thus far has focused on the control matrices $B,b$ (linear) or matrices $B,C,D,b$ (nonlinear), our general approach includes consideration of partitioning the state space into $M\ge1$ distinct regions and similarly partitioning the control matrices into submatrices $\{B_m,b_m\}$ (linear) or submatrices $\{B_m,C_m,D_m,b_m\}$ (nonlinear) such that the submatrices indexed by $m$ are employed when the system is in states composing region $m=1,\ldots,M$.
This will be further discussed in subsequent sections.

W.r.t\ our general control-based RL approach, we establish the following theoretical result on the guaranteed convergence to optimal stabilizing feedback control matrices for the system.
\begin{theorem}
Under the assumption that optimal control matrices exist, the general control-based RL approach above will asymptotically converge to a set of optimal control matrices.
\label{thm:main}
\end{theorem}

In the next section we will consider additional theoretical support for our general control-based RL approach through a general
MDP
framework, involving convergence and optimality both in general and within various control paradigms. Before doing so, however, we briefly discuss an illustrative example.
As one simple generic example of our general control-based RL approach, consider a (unknown) dynamical system model $\dot{x} = f(x)$ and further consider the existence of a linear control model ($B$,$b$) such that $\dot{x} = f(x) + Bx + b$ will solve the problem at hand; analogously consider the existence of a simple nonlinear (quadratic) control model ($B$,$C$,$D$,$b$) such that $\dot{x} = f(x) + Bx + Cx^T Dx + b$ solves the problem.
To illustrate how the system dynamics model tends to have higher complexity and dimensionality than that of the optimal control model within the context of this simple example, consider the Lunar Lander problem of Section~\ref{sec:results} where the system is nonlinear and the state space is of dimension $6$, implying that each linearized vector field is of size $6 \times 6$, has $36$ elements, and depends on the current state.
This is in comparison with only $2$ control dimensions
(left/right and vertical),
and thus the linear control matrix $B$ and vector $b$ (which do not depend on the state) are of size $6 \times 2$ and $1\times 2$, respectively, having a total of only $14$ elements.
Analogously,
the simple nonlinear (quadratic) control
uses the scalar $x_1^2$ in addition to the $6$-dimensional state, which
results in a $7 \times 2$ matrix and a $1\times 2$ vector, totaling $16$ elements.
Moreover, by exploiting additional knowledge, it can be possible to further restrict the degrees of freedom of the optimal control model.

\section{Markov Decision Process Framework}
\label{sec:MDPframework}
We now turn to present a general MDP framework that provides additional theoretical support for our control-based RL approach and, in doing so, extends the classical Bellman operator and optimality criteria by generalizing the definition and scope of a policy for any state.
This includes consideration of convergence and optimality, both in general and within various control paradigms such as linear and piecewise linear control policies, and convergence of $Q$-learning within the context of our framework.

Consider a discrete-space, discrete-time discounted MDP denoted by $(\setX, \setA, \pr, \rewardR, \gamma)$, where $\setX$, $\setA$, $\pr$, $\rewardR$ and $\gamma \in [0,1)$ are as defined above.
Let $\setQ(\setX \times \setA)$ denote the space of bounded real-valued functions over $\setX \times \setA$ with supremum norm.
Furthermore, for any $Q \in \setQ$, define $V(x) := \max_a Q(x,a)$ with the same definition used for variants such as $\hat{Q} \in \setQ$ and $\hat{V}(x)$.
For the state $x_t$ at time $t\in\Ints$ in which action $a$ is taken, i.e., $(x_t,a) \in \setX \times \setA$, denote by $\pr(\cdot | x_t,a)$ the conditional transition probability for the next state $x_{t+1}$ and precisely define $\ex_{\pr} := \ex_{x_{t+1} \sim \pr(\cdot | x_t,a)}$ to be the expectation w.r.t.\ $\pr(\cdot | x_t,a)$.

We next introduce two key ideas to this standard MDP framework, one extending the policy $\pi : \setX \rightarrow \setA$ associated with the Bellman operator to a family of such policies and the other extending the domain of these policies from a single state $x$ to a neighborhood of states associated with $x$.
More formally, define $\cF$ to be a family of functions $f:\setX\rightarrow \setA$ between the set of states and the set of actions. The family $\cF$ can be considered as the set of policies for the MDP, among which we will search to find the best policies.
It is evident that, in the case $\cF$ contains the functions $f(x) = \argmax_{a\in \setA} q(x,a)$ for all $q(x,a) \in \setQ(\setX\times \setA)$, then our introduction of the family $\cF$ continues to correspond to the standard MDP framework where the Bellman equation and operator are instrumental in calculating the optimal policies. 
Furthermore, for each $Q$-function $q(x,a) \in \setQ(\setX\times \setA)$,  we define the function $q^\cF:\setX\times \cF \rightarrow \Reals$ as $q^\cF(x,C) = q(x,C(x))$. It is readily apparent that $q^\cF(x,C)\in \setQ(\setX\times \cF)$.

Turning to the second key aspect of our general MDP framework, let $\nu:\setX\rightarrow 2^{\setX}$ be a function mapping a state to a set of states with $x\in\nu(x)$ that defines a neighborhood of interest for each state $x\in\setX$. Further define a generic operator $\NeighborOp_\nu:\setQ (\setX\times \cF) \rightarrow \setQ (\setX\times \cF)$ that produces a new $q^\cF$ function value at any point $(x,C)$ from the neighborhood set $\{q^\cF(z,C),  z\in\nu(x)\}$. When $\nu(x) = \{x\}$, we fix $\NeighborOp_\nu$ such that $\NeighborOp_\nu q^\cF(x,C) = q^\cF(x,C)$. 
For general
functions $\nu$, 
examples based on $\NeighborOp_\nu$ as the maximum, minimum and average operator respectively render $\NeighborOp_\nu q^\cF(x,C) = \max_{y\in v(x)} q^\cF(y,C)$, $\NeighborOp_\nu q^\cF(x,C) = \min_{y\in v(x)} q^\cF(y,C)$, and $\NeighborOp_\nu q^\cF(x,C) = \int_{v(x)} q^\cF(y,C) d \zeta(y)$ with probability measure $\zeta(y)$ defined on $\nu(x)$; for the special case of $\zeta(y)$ as the point measure at $x$, then 
 $\NeighborOp_\nu q^\cF(x,C) = q^\cF(x,C)$ which is equivalent to the case $\nu(x) = \{x\}$.
By definition, these notions of the neighborhood function $\nu$ and the associated operator $\NeighborOp_\nu$ satisfy the following claim.
\begin{claim}
\label{asm:neighborhood}
For any $x, y \in \setX$, if $\nu(x) =\nu(y)$, then $\NeighborOp_\nu q^\cF(x,C) = \NeighborOp_\nu q^\cF(y,C)$ for any $C\in \cF$. 
\end{claim}

In the remainder of this section, we derive and establish theoretical results within the context of our general MDP framework to support our control-based approach.
We first consider convergence
w.r.t.\ 
both a contraction operator
and
a corresponding form of $Q$-learning, 
followed by 
various aspects of optimality,
and then integration of our control-based approach and our MDP framework.

\subsection{Convergence}
\label{sec:contraction}
We first seek to show that an analog of the Bellman operator within our general MDP framework is a contraction mapping in supremum norm. More specifically, the $Q$-function is a fixed point of a contraction operator $\ContractionOp_\nu$ that is defined, for any function $q^\cF(x,C)\in \setQ(\setX\times \cF)$, to be
$(\ContractionOp_\nu q^\cF)(x,C) =  \sum_{y\in \setX} P_{C(x)}(x,y) [r(x,C(x),y) + \gamma \sup_{D\in\cF} \NeighborOp_\nu q^\cF(y,D) ]$.
Then, for any two functions $q^{\cF}_1(x, C)\in \setQ(\setX\times \cF)$ and $q^\cF_2(x, C)\in \setQ(\setX\times \cF)$, we have
\begin{align*}
 \|\ContractionOp_\nu q^{\cF}_1-\ContractionOp_\nu q^{\cF}_2 \|_{\infty} =&
\sup_{x\in\setX,C\in\cF} \gamma \bigg| \sum_{y\in \setX} P_{C(x)}(x,y) \Big[ \sup_{D_1\in\cF} \NeighborOp_\nu q^{\cF}_1(y,D_1) - \sup_{D_2\in\cF} \NeighborOp_\nu q^{\cF}_2(y,D_2) \Big] \bigg| 
\\
\leq&
\sup_{x\in\setX,C\in\cF} \gamma \sum_{y\in \setX} P_{C(x)}(x,y) \sup_{z\in\setX,D\in\cF} \left| \NeighborOp_\nu q^{\cF}_1(z,D) - \NeighborOp_\nu q^{\cF}_2(z,D) \right| \\
\leq & \sup_{x\in\setX,C\in\cF} \gamma \sum_{y\in \setX} P_{C(x)}(x,y) \| \NeighborOp_\nu q^{\cF}_1 - \NeighborOp_\nu q^{\cF}_2 \|_{\infty}
=
\gamma \|\NeighborOp_\nu q^{\cF}_1 - \NeighborOp_\nu q^{\cF}_2\|_{\infty},
\end{align*}
where the first equality is by definition
and straightforward algebra,
the first inequality is due to multiple applications of the triangle inequality,
and the remaining directly follow for well-behaved operators $\NeighborOp_\nu$.
%
Under appropriate conditions on $\ContractionOp_\nu$ and with $\gamma \in(0,1)$, this establishes that the operator is a contraction in the supremum norm, and therefore $\ContractionOp^t_\nu(q^\cF)$ converges to a unique fixed point of the contraction operator in the limit as $t\rightarrow\infty$, for any function $q^\cF : \setX\times \cF \rightarrow \Reals$.

The above derivation highlights the properties of the generic operator $\NeighborOp_\nu$ that render the desired contraction mapping and convergence of $\ContractionOp_\nu^t(q^\cF)$.
Returning to the examples above,
we observe that, for
 $\NeighborOp_\nu(q^{\cF}(x,C)) = \int_{\nu(x)} q^\cF(y,C) d\mu_x(y)$ with a sub-probability measure $\mu_x$ on $\nu(x)$,
the following inequality holds: 
$\|\NeighborOp_\nu q^{\cF}_1 - \NeighborOp_\nu q^{\cF}_2\|_{\infty} \leq \|\sup_{y \in \nu(x)} \, |q^{\cF}_1(y,C) - q^{\cF}_2(y,C)| \, \|_{\infty}\le\| q^{\cF}_1 - q^{\cF}_2\|_{\infty}$.
For this general $\NeighborOp_\nu$, which includes as special cases the maximum, minimum and average operators,
we have
with $\gamma \in(0,1)$
that
$\ContractionOp_{\nu_t}$ is a contraction and that $\ContractionOp_{\nu_t}^t(q^\cF)$ converges to a unique fixed point
as $t\rightarrow\infty$.
Here $\nu_t$ is the neighborhood function for iteration $t$, where we allow the neighborhood to change over the iterative process.
In all cases above, we observe that
convergence of the contraction operator is to the unique fixed point $q_*^\cF(x,C)$ which satisfies
\begin{align}
\label{eqn:fixed_point}
q_*^\cF(x,C) =  \sum_{y\in \setX} P_{C(x)}(x,y) [r(x,C(x),y) + \gamma \sup_{D\in\cF} \NeighborOp_\nu q_*^\cF(y,D) ] .
\end{align}

We next consider convergence
of the $Q$-learning algorithm within the context of our general MDP framework.
In particular, we focus on the following form of the classical Q-learning update rule \cite{Watk89}:
\begin{equation}
 {q}_{t+1}^{\cF}(x_t,C_t) = {q}_{t}^{\cF}(x_t,C_t) + \alpha_t(x_t,C_t)\left[ r_t + \gamma \sup_{C\in\cF} \NeighborOp_\nu {q}_t^{\cF}(x_{t+1},C) - {q}_t^{\cF}(x_t,C_t)\right], \label{eqn:q-learn-new}
\end{equation}
for $0 < \gamma < 1$ and $0 \leq \alpha_t(x_t,C_t)\leq 1$. Let $C_t$ be a sequence of controllers that covers all state-action pairs and $r_t$ the corresponding reward of applying $C_t$ to state $x_t$. We then have the following result.

\begin{theorem}
\label{thm:QLearning}
Suppose $\NeighborOp_\nu$ is defined as above such that $\ContractionOp_\nu$ is a contraction.
If $\sum_t \alpha_t = \infty$, $\sum_t \alpha_t^2 < \infty$, and $r_t$ are bounded, then
${q}_{t}^{\cF}$ converges to ${q}_*^{\cF}$ as $t\rightarrow\infty$.
\end{theorem}

\subsection{Optimality}
\label{sec:optimality}
We now consider aspects of optimality within our general MDP framework, which first enables us to derive an {\it optimal} policy from $q_*^\cF(x,\cdot)$ as follows:
$D_{\cF,\NeighborOp_\nu} (x) = \argmax_{C\in\cF} \NeighborOp_\nu  q_*^\cF(x,C)$.
%
Define $V^*(x) := \sup_{C\in\cF} \NeighborOp_\nu q^\cF_*(x,C)$.
Then, from \eqref{eqn:fixed_point}, we have
\begin{align}
V^*(x) &= \sup_{C\in\cF} \NeighborOp_\nu q^\cF_*(x,C)
= \sup_{C\in\cF} \NeighborOp_\nu \Big(\sum_{y\in \setX} P_{C(x)}(x,y) [r(x,C(x),y) + \gamma V^*(y)] \Big).\label{eq:V*}
\end{align}

Let us start by considering the optimal policy $D_{\cF,\NeighborOp_\nu} (x)$ when $\nu(x) =\{x\}$, in which case
%
\eqref{eq:V*} becomes
$V^*(x)  = \sup_{C\in\cF} (\sum_{y\in \setX} P_{C(x)}(x,y) [r(x,C(x),y) + \gamma V^*(y)] ).$
This represents an optimality equation for the control policies in the family of functions $\cF$.
We therefore have the following.
%
\begin{theorem}\label{thm:CaseI}
Let $\nu(x)=\{x\}, \forall x\in \setX$. Then, $V^*(x)$ coincides with the value function of the MDP under the family $\cF$ of control policies.
Assuming $\cF$ is sufficiently rich to include a function that assigns an optimal policy when in state $x, \forall x\in\setX$, $V^*$ is the same as the optimal value function for the MDP. Furthermore, when $\cF$ is the family of all (piecewise) linear control policies, $V^*$ represents the best objective function value that can be achieved by (piecewise) linear decision functions.
\end{theorem}

We next consider the case of general neighborhood functions $\nu$ and focus on the maximum operator
$\NeighborOp_\nu q^\cF(x,C) = \max_{y\in v(x)} q^\cF(y,C)$.
%
%
We observe that 
removal of $\NeighborOp_\nu$ in \eqref{eq:V*} decreases the right hand side, and thus it is obviously true that
$V^*(x) \ge \sup_{C\in\cF} \sum_{y\in \setX} P_{C(x)}(x,y) [r(x,C(x),y) + \gamma V^*(y)]$.
This then leads to the following result.
\begin{theorem}\label{thm:CaseII}
Let the neighborhood function $\nu(x), \forall x\in\setX$, be generally defined and let the neighborhood operator be the maximum operator, i.e., $\NeighborOp_\nu q^\cF(x,C) = \max_{y\in v(x)} q^\cF(y,C)$. Then $V^*(x)$ is an upper bound on the optimal value function of the MDP, for all $x\in\setX$. 
\end{theorem}

Let us now turn to consider optimality at a different level within our general MDP framework, where we suppose that the neighborhoods consist of a partitioning of the entire state space $\setX$ into $M$ regions (consistent with Section~\ref{sec:controlRL}), denoted by $\Omega_1, \ldots, \Omega_m$, where
$\nu(x)=\Omega_m, \forall x\in\Omega_m, m=1,2,\ldots, M$.
From Claim \ref{asm:neighborhood}, the right hand side of \eqref{eq:V*} 
has the same value for all states $x$ in the same neighborhood, and therefore we have $V^*(x)=V^*(y), \forall x,y \in \Omega_m, m=1,2,\ldots, M$.
We can write \eqref{eq:V*} as
\begin{align}
\label{eq:HJB_general_aggregated}
V_i & = \sup_{C\in\cF} \NeighborOp_\nu \bigg(\sum_{m=1}^M \Big( \sum_{y\in \Omega_m} P_{C(x)}(x,y) [r(x,C(x),y) + \gamma V_m]\Big) \bigg), \qquad i=1,2,\ldots, M,
\end{align}
where $V_i$ represents the $V^*(x)$ for $x\in \Omega_i$.
These equations can be further simplified depending on the different neighborhood operators, which leads to the following results for optimality and an optimal policy $D_{\cF,\NeighborOp_\nu} (x)$ at the aggregate level in which the policies for state $x$ and state $y$ are the same whenever $x,y\in\Omega_m, m=1,\ldots,M$ (also consistent with Section\ref{sec:controlRL}).

\begin{theorem}\label{thm:CaseIII}
Let the state space $\setX$ be partitioned into disjoint sets $\Omega_m$ such that $\nu(x)=\Omega_m, \forall x\in\Omega_m$, and let $p(x)$ be the function that maps a state to the region (partition) to which it belongs.
For the average neighborhood operator $\NeighborOp_\nu q^\cF(x, C)= \sum_{z\in \Omega_{p(x)}} \zeta(z) q^\cF(z, C)$ with probability vector $\zeta(z)$ satisfying $\sum_{z\in \Omega_m} \zeta(z) =1$, $m=1,2,\ldots, M$, the corresponding version of \eqref{eq:HJB_general_aggregated} is the Bellman equation that produces the optimal value function for the MDP at the aggregate level.
Furthermore, for the maximum neighborhood operator $\NeighborOp_\nu q^\cF(x,C) = \max_{y\in v(x)} q^\cF(y,C)$, the corresponding version of \eqref{eq:HJB_general_aggregated} is the Bellman equation that produces the optimal value function for the MDP at the aggregate level.
\end{theorem}

\subsection{Control-Based Approach and MDP Framework}
\label{sec:control+MDP}
Lastly we consider combinations of our general control-based RL approach together with our general MDP framework.
Theorem~\ref{thm:main} guarantees that our control-based RL approach can asymptotically converge to a set of optimal control matrices, under the assumption of the existence of such optimal control matrices.
This holds for the case of linear or piecewise linear control models $\{(B_m,b_m)\}$ and nonlinear or piecewise nonlinear (quadratic) control models $\{(B_m,C_m,D_m,b_m)\}$.
When these control models contain an optimal policy, then 
Theorems~\ref{thm:CaseI} and \ref{thm:CaseIII} 
provide support for the optimality of the results and Theorem~\ref{thm:main} ensures convergence to an optimal solution.
On the other hand, if the class of control models of interest do not contain any optimal policies, then Theorem~\ref{thm:CaseII} establishes that an asymptotic analysis of these control models can lead to an upper bound on the optimal value function, thus providing theoretical support for such an approach as a good approximation.

Now consider an even tighter integration between our general control-based RL approach and our general MDP framework. Suppose that we start with a coarse partitioning of the state space, say $M=1$ and thus $\nu(x)=\nu(y), \forall x,y\in\setX$. Let us focus on the case of linear control models, with the understanding that more general control models can be employed in an analogous manner. We iteratively follow this instance of our general control-based RL approach from Section~\ref{sec:controlRL} for some number of episodes, either until reaching convergence according to Theorem~\ref{thm:main} with support from Theorem~\ref{thm:CaseIII} or until achieving a desired level of performance relatively close to doing so. Then we increase $M$, partitioning the state space more finely, and once again iteratively follow this instance of our general control-based RL approach with the piecewise linear control models $\{(B_m,b_m)\}$ for some number of episodes along the very same lines.
Once again, our results from Theorems~\ref{thm:main} and \ref{thm:CaseIII} continue to hold and provide theoretical support.
Further proceeding in this manner, with increasing $M$ and decreasing neighborhood sizes, we eventually reach the situation where $\nu(x)=\{x\}, \forall x\in\setX$.  Then, by endowing $\cF$ with a sufficiently rich set of functions, Theorem~\ref{thm:CaseI} together with Theorem~\ref{thm:QLearning} provide the necessary theoretical support that establish the resulting solution to be optimal. The empirical results in the next section provide further support for such an approach.

\section{Experimental Results}
\label{sec:results}
%
We
consider $3$ problems from OpenAI Gym 
\cite{AIgym}~~--~~Lunar Lander, Mountain Car and Cart Pole~~--~~and
present empirical results
that compare the performance of our control-based RL approach (CBRL) against that of the classical $Q$-learning approach with the Bellman operator (QLBO).
The state space of each problem is continuous, which is directly employed in
CBRL.
However, for
QLBO,
the space is discretized to a finite set of states where each dimension is partitioned into equally spaced bins and the number of bins depends on both the problem to be solved and the reference codebase. Specifically, these experiments were executed using the existing code found at \cite{OpenAIcode1,OpenAIcode2}, exactly as is with the default parameter settings.
The exploration and exploitation aspects of
CBRL
include an objective that applies the methods over $K$ episodes with random initial conditions and averages the $K$ rewards; this is then solved using differential evolution \cite{storm:de1997} to find a controller that maximizes the objective function. 
Under the assumption that the system dynamics are unknown and that (piecewise) linear or nonlinear state feedback control is sufficient to solve
the problem,
the goal
becomes learning the parameters of the corresponding optimal control model.
We consider a (unknown) dynamical system model $\dot{x} = f(x, t)$ and a control action $a$ that is a function of
$\nu(x)$
with several simple forms for $a$.
This includes
the control models: ~(i) \emph{Linear}: $a = Bx + b$; ~(ii) \emph{Piecewise Linear}: $a= B_{p(x)}x + b_{p(x)}$ where $p(x)=1,\ldots,M$ is the partition index for state $x$; ~(iii) \emph{Nonlinear}: $a= Bx +b + g(x)$.
%
In summary, our CBRL control models provide significantly better performance in both the training and testing phases for each of the Lunar Lander, Mountain Car and Cart Pole problems considered.

\begin{figure}[htbp]
\vskip -0.5cm
\centering
\subfloat[Average Lunar Lander Score.\label{fig:lunarlander}]{\includegraphics[width=0.33\textwidth]{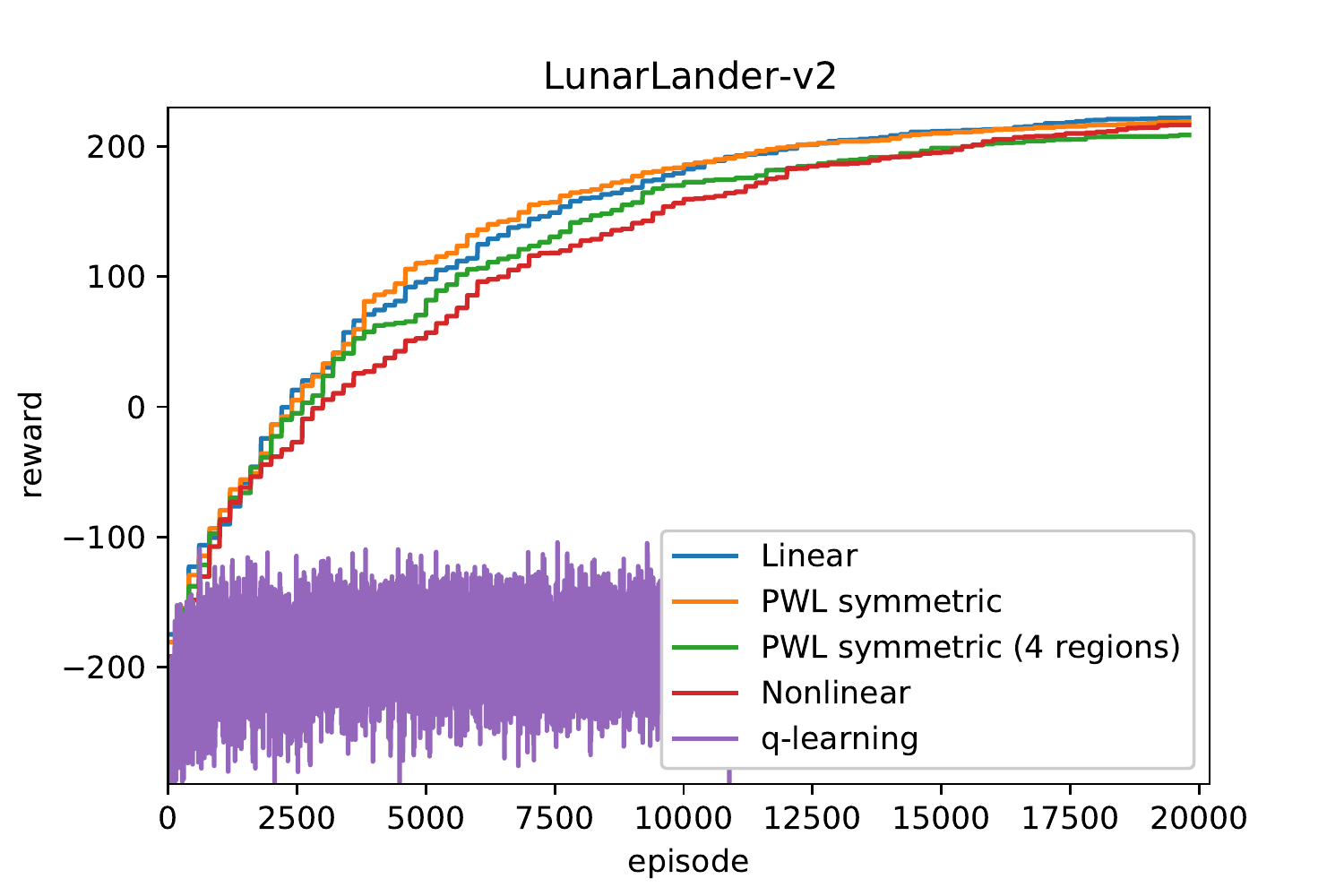}}
\subfloat[Average Mountain Car Score.\label{fig:mountaincar}]{\includegraphics[width=0.33\textwidth]{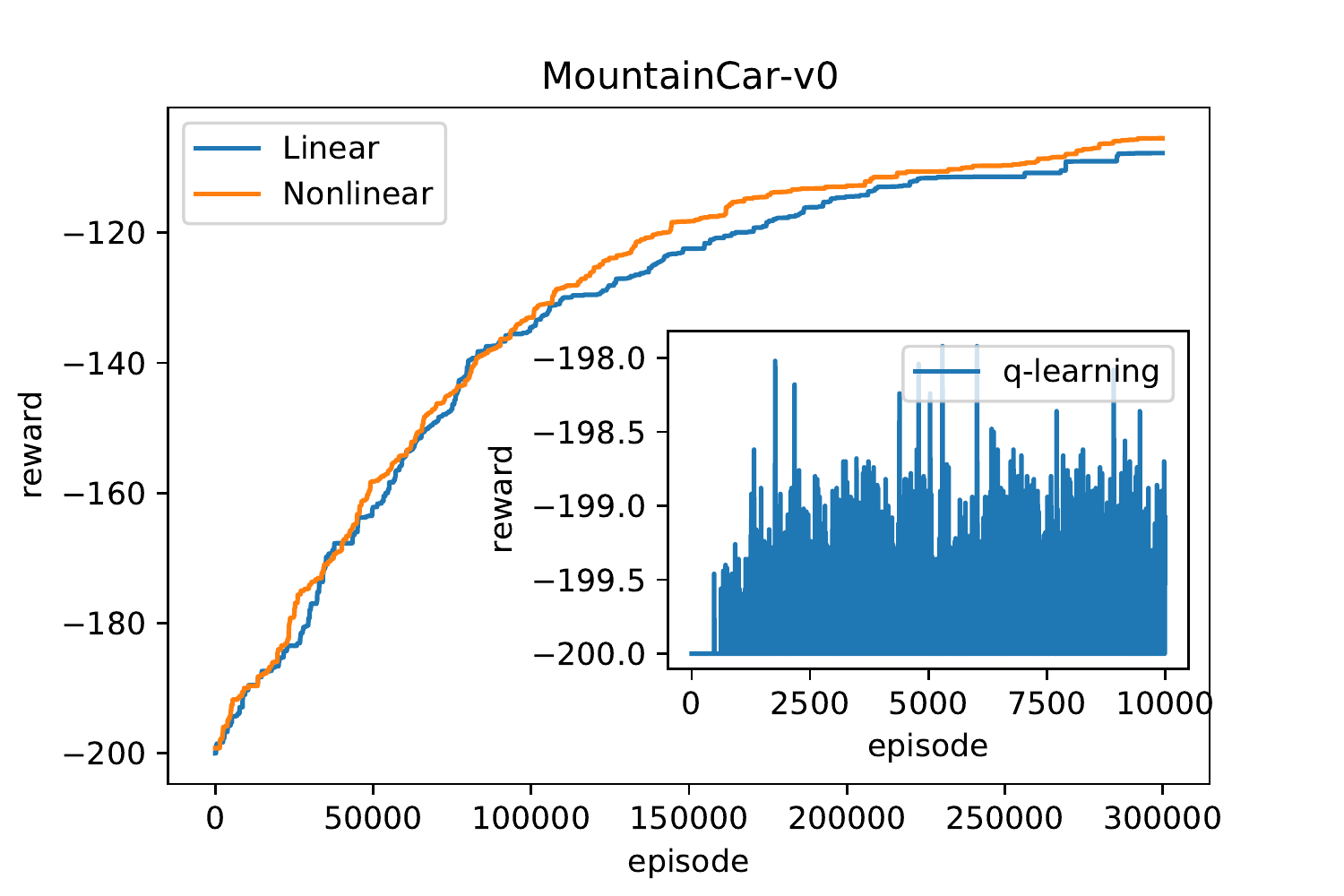}}
\subfloat[Average Cart Pole Score.\label{fig:cartpolelocalsearch}]{\includegraphics[width=0.33\textwidth]{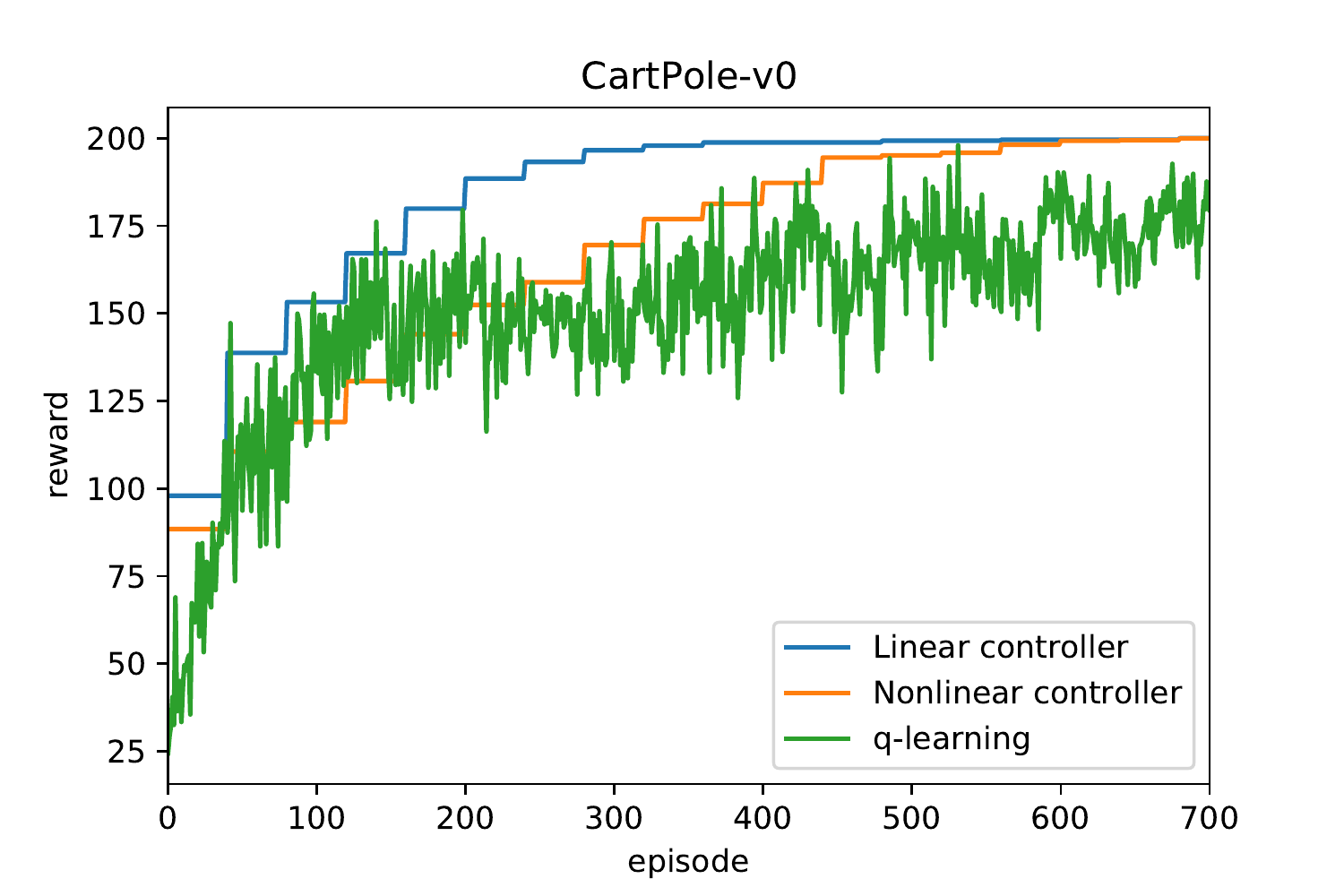}}
\caption{Performance Comparison of Various Instances of Our Control-Based RL Approach and Classical $Q$-Learning with the Bellman Operator, Averaged over $50$ Trials, During Training Phase.}
\label{fig:lunarlander-mountaincar-cartpole}
\vskip -0.5cm
\end{figure}

\paragraph{Lunar Lander.}
This problem is discussed in
\cite{AIgym}.
In each state, characterized by an $8$-dimensional
vector, there are $4$ possible discrete actions (left, right, vertical, no thrusters).
The goal is to maximize the cumulative reward (score) comprising positive points for successful degrees of landing and negative points for fuel usage and crashing.
For the
QLBO experiments,
the $6$ continuous state variables are each discretized into $4$ bins.
For the CBRL experiments, we consider $4$ different simple control models:
Linear;
Piecewise-Linear (PWL) Symmetric with $M=2$, $B_1=B_2$ except for $[B_1]_{*,1}=-[B_2]_{*,1}$, 
where $[B]_{*,1}$ denotes the first column of matrix $B$, and $b_1=b_2$;
Piecewise-Linear (PWL) Symmetric with $M=4$, $B_1=B_4$ except for $[B_1]_{*,1}=-[B_4]_{*,1}$, $b_1=b_4$,
$B_2=B_3$ except for $[B_2]_{*,1}=-[B_3]_{*,1}$, and $b_2=b_3$;
Nonlinear with $g(x)=cx_1^2$ for scalar $x_1^2$.

Figure~\ref{fig:lunarlander} plots the score results, averaged over $50$ experimental trials, as a function of the number of episodes for CBRL under the $4$ simple control models and for QLBO during the training phase.
We observe that QLBO consistently exhibits significantly negative average scores with a trend over many episodes that remains flat.
In strong contrast, our CBRL with each of the simple control models finds control policies that achieve average scores above $+200$ within $20,000$ episodes with trends continuing to slightly improve.
We further observe that all $4$ CBRL result curves are quite similar.



Table \ref{tbl:lunarlander} (Appendix~\ref{app:lunarlander}) presents
the average score over $1000$ episodes across the $50$ trials for CBRL under the $4$ simple control models and for QLBO during the testing phase, together with the corresponding 95\% confidence intervals.
We again observe that the best average scores are by far obtained under the simple CBRL control models, with Linear considerably above $+200$ and QLBO below $-200$,
and that the confidence intervals for all cases are quite small.
We further observe the performance orderings among the simple CBRL control models with Linear providing the best results, followed by Nonlinear, PWL Symmetric ($M=4$), and PWL Symmetric ($M=2$). Note that the Linear and PWL Symmetric ($M=2$) controllers have the same number of parameters and they also provide the highest rewards during training; while the other CBRL controllers eventually approach these levels of reward performance, they do so more slowly due to the increased degrees of freedom (parameters). For the testing results, the CBRL controllers that provide better reward performance also have smaller confidence intervals, suggesting that they are closer to an optimal controller.
Furthermore, the Linear controller provides  the best performance, suggesting that the Nonlinear and PWL controllers require further searching of the decision space in the training phase to identify the corresponding optimal controller for the testing phase.


\paragraph{Mountain Car.}
This problem is
discussed in \cite{mountaincar}.
In each state, characterized by a $2$-dimensional vector, there are $3$ possible actions (forward, backward, neutral).
The goal is to maximize the score representing the negation of the number of time steps needed to solve the problem
(i.e., minimize number of steps to solve) over episodes of up to $200$ steps.
For the QLBO experiments, the state space is discretized into a $40\times 40$ grid.
For the CBRL experiments, we consider $2$ simple control models: Linear; Nonlinear with $g(x)$ as a $2$nd order multivariate polynomial on the state variables.

Figure~\ref{fig:mountaincar} plots the score results, averaged over $50$ experimental trials, as a function of the number of episodes for CBRL under the $2$ simple control models and for QLBO during the training phase.
We observe that the average scores for QLBO continually hover quite close to $-200$, as illustrated in the inset plot for the first $10,000$ episodes, implying that this algorithm is very rarely successful in solving the problem over a very large number of episodes across the $50$ trials.
In strong contrast, the average scores for CBRL with the simple Linear and Nonlinear control models both follow a similar path towards $-110$, which is the point at which the problem is considered to be solved.

Table \ref{tbl:mountaincar} (Appendix~\ref{app:mountaincar}) presents
the average score over $1000$ episodes across the $50$ trials for CBRL under the $2$ simple control models and for QLBO during the testing phase, together with the corresponding 95\% confidence intervals.
We again observe that the best average scores are by far obtained under the simple CBRL control models, with Linear and Nonlinear both solving the problem and QLBO still far away,
and that the confidence intervals for all cases are quite small.
We further observe the performance orderings among the simple CBRL control models with Nonlinear slightly outperforming Linear, but the differences are very small.
While the Nonlinear controller outperforms the Linear controller during both the training and testing phases, both simple CBRL control models perform extremely well, they completely solve the problem, and they significantly outperform QLBO.

\paragraph{Cart Pole.}
%
This problem is discussed in
\cite{cartpole}.
In each state, characterized by a $4$-dimensional vector, there are $2$ possible discrete actions (push left, push right).
The goal is to maximize the score representing the number of steps that the cart pole stays upright before either falling over or going out of bounds.
%
%
For the QLBO experiments, the position and velocity are discretized into $8$ bins whereas the angle and angular velocity are discretized into $10$ bins. %
For the CBRL experiments, we consider $2$ different simple control models: Linear; Nonlinear with $g(x)$ as a $2$nd order multivariate polynomial on the state variables.
With a score of $200$, the problem is considered solved and the simulation ends.
%

Figure~\ref{fig:cartpolelocalsearch} plots the score results, averaged over $50$ experimental trials, as a function of the number of episodes for CBRL under the $2$ simple control models and for QLBO during the training phase.
We observe that CBRL, with both the simple Linear and Nonlinear control models, quickly finds an optimal control policy that solves the problem within a few hundred episodes, whereas QLBO continues to oscillate well below the maximal score of $200$ even for this relatively simple problem.

Table \ref{tbl:cartpole} (Appendix~\ref{app:cartpole}) presents
the average score over $1000$ episodes across the $50$ trials for CBRL under the $2$ simple control models and for QLBO during the testing phase, together with the corresponding 95\% confidence intervals.
We again observe that the best average scores are by far obtained under the simple CBRL control models, with Linear and Nonlinear both solving the problem while having confidence intervals of essentially zero and with QLBO providing poorer performance and still a considerable distance away from optimal while having a very small confidence interval. In addition, the Linear controller with less degrees of freedom reaches higher scores more quickly than the Nonlinear controller, but both controllers reach a score of $200$ after a few hundred episodes.

\section*{Broader Impact}
Our contributions are in the areas of reinforcement learning (RL) and Markov decision processes (MDP), and as such have the potential of impacting a very broad spectrum of applications in business, engineering and science.
The great success of RL to solve a wide variety of learning and decision problems can be further enhanced through the approach and results of the paper.
Since our approach and results improve the sample complexity problem of RL, we believe there is the potential of enabling more applications to learn faster and with less data.
In addition, we plan to include the main approaches and results of this paper as part of the curriculum in graduate courses.

\bibliography{main}
\bibliographystyle{abbrvnat}

\clearpage
\newpage
\newpage
\newpage

\appendix

\section{Generic Algorithm for General Control-Based Reinforcement Learning}
\label{app:alg}
Our general control-based RL approach can be summarized by the following generic simplified algorithm, using the notation described in Sections~\ref{sec:controlRL} and \ref{sec:MDPframework}.
%
\begin{algorithm}[htbp] 
\caption{General Control-Based RL Method}
	\DontPrintSemicolon
	\SetKwFunction{FindNextControl}{Find\_Next\_Control\_Model}
	\SetKwFunction{UpdateTable}{Update\_Table\_and\_Model}
	\SetKwFunction{RunEpisode}{Run\_Episode}
	\SetKwInOut{Input}{Input}\SetKwInOut{Output}{Output}
	\Input{  Initial control matrices $(\bB_0, \bC_0, \bD_0, \bb_0)$. }
	\Output{ Set of best control matrices $C^*_p = (\bB^*, \bC^*, \bD^*, \bb^*)\in \cF$, Corresponding $q$-table $q_*^\cF(\cdot,\cdot)$.}
	\;
	
	Initialize $C_p = (\bB_0, \bC_0, \bD_0, \bb_0)$, Initialize $q$-table $q_*^\cF(\cdot,\cdot)$ to be empty. \;
	\For {episode $e \in [1, \ldots\,)$}
		{ $(r(t))_{t\in[T_e]}, \; q^\cF(\cdot,\cdot) \, = \,$ \RunEpisode($C_p, q^\cF(\cdot,\cdot)$) \;
		$C_p \, = \, $ \FindNextControl ($C_p$, $q^\cF(\cdot,\cdot)$)\;
		\If {tolerance satisfied for $\{ (r(t))_{t\in[T_e]}, C_p, q^\cF(\cdot,\cdot)\}$ }{\Return{$C_p$, $q^\cF(\cdot,\cdot)$}}
		}
	\;
	\label{step:id_feat1}\Fn{\RunEpisode}{
	\KwData{ Current control matrices $C_p$ for episode, Current $q$-table $q^\cF(\cdot,\cdot)$.}
	\KwResult{ Reward sequence $(r(t))_{t\in [T]}$ from episode based on $C_p$, Updated $q$-table $q^{\cF}(\cdot,\cdot)$.}
 	\For {$t \in \{ 1,\ldots,T \}$}
 	    {
        $(x(t),a(t),r(t)) \, =\,$ Execute system (or simulation thereof) using $a(t) = C_p(x(t))$ to obtain corresponding sample measurement for state, action, reward tuple.\;
        Update $q^\cF(\cdot,\cdot)$ with $(x(t),a(t),r(t))$ 
	    }
    \Return{$(r(t))_{t\in[T]}$, $q^\cF(\cdot,\cdot)$}
	}
	\;
	\label{step:id_feat3}\Fn{\FindNextControl}{
	\KwData{ Current best control matrices $C_p$, Current $q$-table $q^\cF(\cdot,\cdot)$. }
	\KwResult{ Next control matrices $C_p$. }
	Set $C_p$ based on black-box optimization (e.g., local search, differential evolution, Bayesian optimization, genetic algorithms) and $q^\cF(\cdot,\cdot)$\;
	\Return{$C_p$}
	}
\label{alg:find_control}
\end{algorithm}

\section{Control-Based Reinforcement Learning}
\label{app:controlRL}
In this section we present additional technical details and results related to Section~\ref{sec:controlRL}.

\subsection{Proof of Theorem~\ref{thm:main}}
\label{app:thm}
%
First, consider the dynamics of a system given by $$ \dot{\bx} = f(\bx,t)$$ for the continuous-time case or by $$\bx_{n+1} = f(\bx_n,n)$$ for the discrete-time case.
Suppose $f$ is Lipschitz continuous with Lipschitz constant $L$ and the goal of the task at hand is to ensure that the trajectory of the system converges toward an equilibrium point of the system $\bx_0$.
Lyapunov stability analysis and known results from time-varying linear systems \cite{ilchmann87} show that $$\bu = \bB\bx$$ is a linear stabilizing feedback provided that
$$Df(\bx_0,t) + \bB$$ has eigenvalues with negative real parts bounded away from the origin and slowly varying, where $Df(\cdot,\cdot)$ is the Jacobian matrix of $f(\cdot,\cdot)$.
In particular, $-\alpha \bI$ is a stabilizing feedback matrix where $\alpha > L$.

This then implies that an appropriate search on $\alpha$ will result in a stabilizing feedback.
Such an optimal search can be performed to find feedback matrices that satisfy bounds, such as spectral bounds, using both gradient-based and gradient-free methods, such as local search,  differential evolution, Bayesian optimization and genetic algorithms.
The eigenvalue conditions show that a feedback matrix within a small ball around $B$ will still stabilize the system, and therefore such instances of our general control-based RL approach will guarantee to find a solution asymptotically given sufficient time to search the space, analogous to the guaranteed asymptotic convergence under the Bellman equation.

\section{Markov Decision Process Framework}
\label{app:MDPframework}
In this section we present additional technical details and results related to Section~\ref{sec:MDPframework}.

\subsection{Convergence}
\label{app:contraction}
We first seek to show that an analog of the Bellman operator within our general MDP framework is a contraction mapping in supremum norm. More specifically, the $Q$-function is the fixed point of a contraction operator $\ContractionOp_\nu$ that is defined, for any function $q^\cF(x,C)\in \setQ(\setX\times \cF)$, to be
\[ (\ContractionOp_\nu q^\cF)(x,C) =  \sum_{y\in \setX} P_{C(x)}(x,y) [r(x,C(x),y) + \gamma \sup_{D\in\cF} \NeighborOp_\nu q^\cF(y,D) ] .\]
Then, for any two functions
$$q^{\cF}_1(x, C)\in \setQ(\setX\times \cF) \qquad \mbox{ and } \qquad q^\cF_2(x, C)\in \setQ(\setX\times \cF) , $$
we have
\begin{align*}
 \|\ContractionOp_\nu q^{\cF}_1-\ContractionOp_\nu q^{\cF}_2 \|_{\infty} =& \sup_{x\in\setX,C\in\cF} \left| \sum_{y\in \setX} P_{C(x)}(x,y) \Big[r(x,C(x),y) + \gamma \sup_{D_1\in\cF} \NeighborOp_\nu q^{\cF}_1(y,D_1) \right. \\  & \left. \qquad\qquad\qquad\qquad\qquad\quad - r(x,C(x),y) - \gamma \sup_{D_2\in\cF} \NeighborOp_\nu q^{\cF}_2(y,D_2) \Big] \right|  
 \\
=& \sup_{x\in\setX,C\in\cF} \gamma \left| \sum_{y\in \setX} P_{C(x)}(x,y) \Big[ \sup_{D_1\in\cF} \NeighborOp_\nu q^{\cF}_1(y,D_1) - \sup_{D_2\in\cF} \NeighborOp_\nu q^{\cF}_2(y,D_2) \Big] \right| 
\\
\leq& \sup_{x\in\setX,C\in\cF} \gamma \sum_{y\in \setX} P_{C(x)}(x,y) \left| \sup_{D_1\in\cF} \NeighborOp_\nu q^{\cF}_1(y,D_1)] - \sup_{D_2\in\cF} \NeighborOp_\nu q^{\cF}_2(y,D_2) \right| 
\\
\leq& \sup_{x\in\setX,C\in\cF} \gamma \sum_{y\in \setX} P_{C(x)}(x,y) \sup_{z\in\setX,D\in\cF} \left| \NeighborOp_\nu q^{\cF}_1(z,D) - \NeighborOp_\nu q^{\cF}_2(z,D) \right| \\
\leq & \sup_{x\in\setX,C\in\cF} \gamma \sum_{y\in \setX} P_{C(x)}(x,y) \| \NeighborOp_\nu q^{\cF}_1 - \NeighborOp_\nu q^{\cF}_2 \|_{\infty} \\
= & \; \gamma \|\NeighborOp_\nu q^{\cF}_1 - \NeighborOp_\nu q^{\cF}_2\|_{\infty},
\end{align*}
where the first equality is by definition, the second equality follows from straightforward algebra, the first and second inequalities are due to the triangle inequality,
and the remaining directly follow for well-behaved operators $\NeighborOp_\nu$.
Under appropriate conditions on $\ContractionOp_\nu$ and with $\gamma \in(0,1)$, this establishes that the operator is a contraction in the supremum norm, and therefore $\ContractionOp^t_\nu(q^\cF)$ converges to a unique fixed point of the contraction operator in the limit as $t\rightarrow\infty$, for any function $q^\cF : \setX\times \cF \rightarrow \Reals$.

The above derivation highlights the properties of the generic operator $\NeighborOp_\nu$ that render the desired contraction mapping and convergence of $\ContractionOp_\nu^t(q^\cF)$.
Returning to the examples above,
we observe that, for 
 $$\NeighborOp_\nu(q^{\cF}(x,C)) = \int_{\nu(x)} q^\cF(y,C) d\mu_x(y)$$ with a sub-probability measure $\mu_x$ on $\nu(x)$,
the following inequality holds: 
$$\|\NeighborOp_\nu q^{\cF}_1 - \NeighborOp_\nu q^{\cF}_2\|_{\infty} \leq \|\sup_{y \in \nu(x)} \, |q^{\cF}_1(y,C) - q^{\cF}_2(y,C)| \, \|_{\infty}\le\| q^{\cF}_1 - q^{\cF}_2\|_{\infty} .$$
For this general $\NeighborOp_\nu$, which includes as special cases the maximum, minimum and average operators,
we have
with $\gamma \in(0,1)$
that
$\ContractionOp_{\nu_t}$ is a contraction and that $\ContractionOp_{\nu_t}^t(q^\cF)$ converges to a unique fixed point
in the limit
as $t\rightarrow\infty$.
Here $\nu_t$ is the neighborhood function for iteration $t$, where we allow the neighborhood to change over the iterative process.
In all cases above, we observe that
the fixed point $q_*^\cF(x,C)$ of the contraction operator satisfies
\begin{align}
\label{eqn:app:fixed_point}
q_*^\cF(x,C) =  \sum_{y\in \setX} P_{C(x)}(x,y) [r(x,C(x),y) + \gamma \sup_{D\in\cF} \NeighborOp_\nu q_*^\cF(y,D) ] .
\end{align}

We next consider convergence
of a form
of the $Q$-learning algorithm within the context of our general MDP framework.
In particular, we focus on the following form of the classical Q-learning update rule \cite{Watk89}:
\begin{equation}
 {q}_{t+1}^{\cF}(x_t,C_t) = {q}_{t}^{\cF}(x_t,C_t) + \alpha_t(x_t,C_t)\left[ r_t + \gamma \sup_{C\in\cF} \NeighborOp_\nu {q}_t^{\cF}(x_{t+1},C) - {q}_t^{\cF}(x_t,C_t)\right], \label{eqn:app:q-learn-new}
\end{equation}
for $0 < \gamma < 1$ and $0 \leq \alpha_t(x_t,C_t)\leq 1$. Let $C_t$ be a sequence of controllers that covers all state-action pairs and $r_t$ the corresponding reward of applying $C_t$ to state $x_t$. We then have the result in Theorem~\ref{thm:QLearning} whose proof is given in the next section.

\subsection{Proof of Theorem \ref{thm:QLearning}} \label{sec:proof-q-learning}
The proof is somewhat similar to the well-known proof of the convergence of the classical $Q$-learning update (see, e.g,, \cite{jaakkola:1994,melo:2001}) within the context of our general MDP framework.  We first rewrite \eqref{eqn:app:q-learn-new} as a convex combination of $${q}_{t}^{\cF}(x_t,C_t) \qquad\qquad \mbox{and} \qquad\qquad r_t + \gamma \sup_{C\in\cF} \NeighborOp_\nu {q}_t^{\cF}(x_{t+1},C) ,$$
where
$$ {q}_{t+1}^{\cF}(x_t,C_t) = \left(1-\alpha_t(x_t,C_t)\right){q}_{t}^{\cF}(x_t,C_t) + \alpha_t(x_t,C_t)\left[ r_t + \gamma \sup_{C\in\cF} \NeighborOp_\nu {q}_t^{\cF}(x_{t+1},C) \right] .$$
Define the difference between $q_t^{\cF}$ and $q_*^{\cF}$ as $\Delta_t := q_t^{\cF} - q_*^{\cF}$, which satisfies
$$ \Delta_{t+1}(x_t,C_t) = \left(1-\alpha_t(x_t,C_t)\right){\Delta}_{t}(x_t,C_t) + \alpha_t(x_t,C_t)\left[ r_t + \gamma \sup_{C\in\cF} \NeighborOp_\nu {q}_t^{\cF}(x_{t+1},C) - q_*^{\cF}\right] .$$

Further define
$$ F_t(x,C) := r(x,C(x),X(x,C)) + \gamma \max_{D\in\cF} q_t^{\cF}(y,D) - q_*^{\cF}(x,C) , $$
where $X(x,C)$ is a randomly sampled state from the MDP under an initial state $x$ and policy $C$, 
from which we obtain
\begin{align*}
|\ex[F_t(x,C)|P_t]| & = \left|\sum_{y\in \setX} P(x,C(x),y)\left[r(x,C(x),y) +\gamma \max_{D\in\cF}q_t^{\cF}(y,D) - q_*^{\cF}(x,C)\right]\right| \\
& =  |\ContractionOp_{\nu}q_t^{\cF}(x,C) - q_*^{\cF}(x,C)| = |\ContractionOp_{\nu}q_t^{\cF}(x,C) - \ContractionOp_{\mu}q_*^{\cF}(x,C)| \leq \gamma \|\Delta_t\|_{\infty} .
\end{align*}
where $P_t = \{\Delta_t, \Delta_{t-1}, \cdots, F_{t-1},\cdots, \alpha_{t-1},\cdots \}$ stands for the past state and parameters \cite{jaakkola:1994}. Similarly, we have
\begin{align*}
    \var[F_t(x,C)|P_t] & = \ex\left[\left(r(x,C,X(x,C))+\gamma\max_{D\in\cF}q_t^\cF(y,D) - \ContractionOp_{\nu}q_t^{\cF}(x,C)\right)^2\right] \\
    & = \var \left[\left.r(x,C,X(x,C))+\gamma\max_{D\in\cF}q_t^{\cF}(y,D)\right|P_t\right] \leq  \kappa(1+\|\Delta_t\|_{\infty}^2) ,
\end{align*}
for some $\kappa > 0$.
The desired result then follows from Theorem 1 in \cite{jaakkola:1994}.

\subsection{Optimality}
\label{app:optimality}
We now consider aspects of optimality within our general MDP framework, which first enables us to derive an {\it optimal} policy from $q_*^\cF(x,\cdot)$ as follows:
\begin{equation} \label{eqn:optimal_policy}
D_{\cF,\NeighborOp_\nu} (x) = \argmax_{C\in\cF} \NeighborOp_\nu  q_*^\cF(x,C) .
\end{equation}
Define $V^*(x) := \sup_{C\in\cF} \NeighborOp_\nu q^\cF_*(x,C)$.
Then, from \eqref{eqn:app:fixed_point}, we have
\begin{align}
V^*(x) &= \sup_{C\in\cF} \NeighborOp_\nu q^\cF_*(x,C) \nonumber\\ &
 = \sup_{C\in\cF} \NeighborOp_\nu \left(\sum_{y\in \setX} P_{C(x)}(x,y) [r(x,C(x),y) + \gamma \sup_{C'\in\cF} \NeighborOp_\nu q^\cF_*(y,C') ] \right)
 \nonumber\\& 
= \sup_{C\in\cF} \NeighborOp_\nu \Big(\sum_{y\in \setX} P_{C(x)}(x,y) [r(x,C(x),y) + \gamma V^*(y)] \Big).\label{eq:app:V*}
\end{align}

Let us start by considering the optimal policy $D_{\cF,\NeighborOp_\nu} (x)$ in \eqref{eqn:optimal_policy} when $\nu(x) =\{x\}$, in which case \eqref{eq:app:V*} becomes
\begin{align}
V^*(x)  = \sup_{C\in\cF} \left(\sum_{y\in \setX} P_{C(x)}(x,y) [r(x,C(x),y) + \gamma V^*(y)] \right).
\label{eq:app:Bellman:thm3.2}
\end{align}
This represents an optimality equation for the control policies in the family of functions $\cF$.
We therefore have the result in Theorem~\ref{thm:CaseI} whose proof is given in Section~\ref{app:thm:CaseI}.

We next consider the case of general neighborhood functions $\nu$ and focus on the maximum operator
$\NeighborOp_\nu q^\cF(x,C) = \max_{y\in v(x)} q^\cF(y,C)$.
In this case, 
we observe that 
removal of $\NeighborOp_\nu$ in \eqref{eq:app:V*} decreases the right hand side, and thus it is obviously true that
\begin{align}
\label{eqn:lower_bound}
V^*(x) &\ge \sup_{C\in\cF} \sum_{y\in \setX} P_{C(x)}(x,y) [r(x,C(x),y) + \gamma V^*(y)].
\end{align}
This then leads to the result in Theorem~\ref{thm:CaseII} whose proof is given in Section~\ref{app:thm:CaseII}.

Let us now turn to consider optimality at a different level within our general MDP framework, where we suppose that the neighborhoods consist of a partitioning of the entire state space $\setX$ into $M$ regions (consistent with Section~\ref{sec:controlRL}), denoted by $\Omega_1, \ldots, \Omega_m$, with $\nu(x)=\Omega_m, \forall x\in\Omega_m, m=1,2,\ldots, M$.
From Claim~\ref{asm:neighborhood}, the right hand side of \eqref{eq:app:V*} has the same value for all states $x$ in the same neighborhood, and therefore we have $V^*(x)=V^*(y), \forall x,y \in \Omega_m, m=1,2,\ldots, M$.
We can express \eqref{eq:app:V*} as
\begin{align*}
V_i & = \sup_{C\in\cF} \NeighborOp_\nu \left(\sum_{y\in \setX} P_{C(z)}(z,y) [r(z,C(z),y) + \gamma V_{p(y)}] \right), \qquad i=1,2,\ldots, M,
\end{align*}
where $V_i$ represents the $V^*(x)$ for $x\in \Omega_i$ and $p(y)$ is the function that maps a state to the region (partition) to which it belongs.
This can be equivalently written as 
\begin{align}
\label{eq:app:HJB_general_aggregated}
V_i & = \sup_{C\in\cF} \NeighborOp_\nu \left(\sum_{m=1}^M \left( \sum_{y\in \Omega_m} P_{C(x)}(x,y) [r(x,C(x),y) + \gamma V_m]\right) \right), \qquad i=1,2,\ldots, M.
\end{align}
These equations can be further simplified depending on the different neighborhood operators, which leads to the results in Theorem~\ref{thm:CaseIII} for optimality and an optimal policy $D_{\cF,\NeighborOp_\nu} (x)$ at the aggregate level where the policies for state $x$ and state $y$ are the same whenever $x,y\in\Omega_m, m=1,\ldots,M$ (also consistent with Section\ref{sec:controlRL}).
The proof is provided in Section~\ref{app:thm:CaseIII}.

\subsection{Proof of Theorem~\ref{thm:CaseI}}
\label{app:thm:CaseI}
%
%
With $\nu(x)=\{x\}, \forall x\in \setX$, we know from \eqref{eqn:optimal_policy} and \eqref{eq:app:V*} that the value function $V^*(x)$ satisfies  \eqref{eq:app:Bellman:thm3.2}, which represents an optimality equation for the control policies in the family of functions $\cF$. Hence, $V^*(x)$ coincides with the value function of the MDP under the family $\cF$ of control policies.
Under the assumption that $\cF$ is sufficiently rich to include a function that assigns optimal policy when in state $x, \forall x\in\setX$ (such as the one determined by the Bellman operator but not restricted to this specific optimal policy), then this value function $V^*$ is the same as the optimal value function for the MDP.
On the other hand, under the assumption that $\cF$ is the family of all (piecewise) linear control policies, then this value function $V^*$ represents the best objective function value that can be achieved by (piecewise) linear decision functions.

\subsection{Proof of Theorem~\ref{thm:CaseII}}
\label{app:thm:CaseII}
%
For an MDP within the context of our framework, the optimal value function that satisfies
\begin{align*}
v(x) &= \sup_{C\in\cF} \sum_{y\in \setX} P_{C(x)}(x,y) [r(x,C(x),y) + \gamma v(y)]
\end{align*}
is well known (see, e.g.,~\cite[Proposition 5.1, Chapter II]{rossSDP})
to be the minimum value of all the functions $v'(x)$ that satisfy
\begin{align*}
v'(x) &\ge \sup_{C\in\cF} \sum_{y\in \setX} P_{C(x)}(x,y) [r(x,C(x),y) + \gamma v'(y)].
\end{align*}
Hence, \eqref{eqn:lower_bound} ensures that $V^*(x)$ is an upper bound on the optimal value function.

\subsection{Proof of Theorem~\ref{thm:CaseIII}}
\label{app:thm:CaseIII}
%
%
With the state space $\setX$ partitioned into disjoint sets $\Omega_m$ such that $\nu(x)=\Omega_m, \forall x\in\Omega_m$, $m=1,\ldots,M$, let us first consider the average neighborhood operator $$\NeighborOp_\nu q^\cF(x, C)= \sum_{z\in \Omega_{p(x)}} \zeta(z) q^\cF(z, C)$$ where the probability vector $\zeta(z)$ satisfies $\sum_{z\in \Omega_m} \zeta(z) =1$, for all $m=1,2,\ldots, M$.
We then have \eqref{eq:app:HJB_general_aggregated} taking the form
\begin{align*}
V_i & = \sup_{C\in\cF} \sum_{z\in \Omega_i} \zeta(z) \left(\sum_{m=1}^M \left( \sum_{y\in \Omega_m} P_{C(z)}(z,y) [r(z,C(z),y) + \gamma V_m]\right) \right), \forall i=1,2,\ldots, M,
\end{align*}
and therefore
\begin{align}
\label{eqn:HJB_aggregated}
V_i & = \sup_{C\in\cF}\sum_{m=1}^M \left(\sum_{z\in \Omega_i} \sum_{y\in \Omega_m} P_{C(z)}(z,y) \zeta(z) r(z,C(z),y)  + \sum_{z\in \Omega_i} \sum_{y\in \Omega_m}P_{C(z)}(z,y)\zeta(z)\gamma V_m]\right) .
\end{align}
Now consider a new MDP with states $1,2, \ldots, M$, action space $\cF$, transition probabilities
$$P_{C}(i,j) :=\sum_{z\in \Omega_i} \sum_{y\in \Omega_j}P_{C(z)}(z,y)\zeta(z) , $$
and rewards
\begin{align*}
r(i,C, j):= \frac{1}{P_{C}(i,j)}\sum_{z\in \Omega_i} \sum_{y\in \Omega_j} P_{C(z)}(z,y) \zeta(z) r(z,C(z),y).
\end{align*}
Hence \eqref{eqn:HJB_aggregated} is the Bellman equation that produces the optimal value function for this new MDP, which corresponds to the original MDP at the aggregate level.

Under the same state space partitioning, let us next consider the maximum neighborhood operator
$$\NeighborOp_\nu q^\cF(x,C) = \max_{y\in v(x)} q^\cF(y,C) . $$ 
We then have \eqref{eq:app:HJB_general_aggregated} taking the form
\begin{align*}
V_i & = \sup_{C\in\cF} \sup_{z\in \Omega_i} \left(\sum_{m=1}^M \left( \sum_{y\in \Omega_m} P_{C(z)}(z,y) [r(z,C(z),y) + \gamma V_m]\right) \right) ,  \forall i=1,2,\ldots, M,
\end{align*}
and therefore
\begin{align}
\label{eqn:HJB_aggregated_max}
V_i & = \sup_{C\in\cF, z\in \Omega_i}
\left(\sum_{m=1}^M \left( \sum_{y\in \Omega_m} P_{C(z)}(z,y) r(z,C(z),y) + \gamma \left(\sum_{y\in \Omega_m} P_{C(z)}(z,y)\right)V_m\right) \right).
\end{align}
Once again, consider a new MDP with states $1,2, \ldots, M$, action sets $\{\Omega_i, \cF\}_i$, transition probabilities
$$P_{C,z}(i,j) :=  \sum_{y\in \Omega_j} P_{C(z)}(z,y) ,$$ and rewards
\begin{align*}
r(i,(C,z),j) := \sum_{y\in \Omega_j} P_{C(z)}(z,y) r(z,C(z),y),
\end{align*}
where $z\in \Omega_i$ in both the transition probability and reward definitions.
Hence \eqref{eqn:HJB_aggregated_max} is the Bellman equation that produces the optimal value function for this new MDP, which corresponds to the original MDP at the aggregate level.

\section{Experimental Results}
\label{app:results}
In this section we present additional material and results that complement those in the main paper.

We consider $3$ problems from OpenAI Gym~\cite{AIgym}~~--~~namely
Lunar Lander, Mountain Car and Cart Pole~~--~~and we present empirical results for each problem that compare the performance of our control-based RL approach against that of the classical $Q$-learning approach with the Bellman operator.
The state space of each problem is continuous, which is directly employed in our control-based RL approach.
However, for classical $Q$-learning with the Bellman operator, the space is discretized to a finite set of states where each dimension is partitioned into equally spaced bins and the number of bins depends on both the problem to be solved and the reference codebase. 
The exploration and exploitation aspects of our control-based RL approach include an objective that applies the methods over $K$ episodes with random initial conditions and averages the $K$ rewards; this is then solved using differential evolution \cite{storm:de1997} to find a controller that maximizes the objective function. 
Under the assumption that the system dynamics are unknown and that (piecewise) linear or nonlinear state feedback control is sufficient to solve the problem at hand, the goal then becomes learning the parameters of the corresponding optimal control model.
We consider a (unknown) dynamical system model $\dot{x} = f(x, t)$ and a control action $a$ that is a function of the neighborhood $\nu(x)$ of the state $x$ with several simple forms for $a$.
This includes (variants of) the control models:
\begin{enumerate}
    \item \emph{Linear}: $a = Bx + b$;
    \item \emph{Piecewise Linear}: $a= B_{p(x)}x + b_{p(x)}$ where $p(x)=1,\ldots,M$ is the partition index for state $x$;
    \item \emph{Nonlinear}: $a= Bx +b + g(x)$ where $g(x)$ is a general function of $x$.
\end{enumerate}

Multiple experimental trials are run for each problem
from the OpenAI Gym.
For each problem,
our experimental results under the classical $Q$-learning approach with the Bellman operator were obtained using the existing code found at \cite{OpenAIcode1,OpenAIcode2} exactly as is with the default parameter settings. This includes use of a standard, yet generic, form for the $Q$-learning algorithms.  Specifically, for all $Q_0 \in \setQ$, $x \in \setX$, $a \in \setA$ and the Bellman operator $\BellmanOp$, the sequence of action-value $Q$-functions is updated based on the following generic rule:
\begin{equation}
Q_{k+1}(x,a) = (1-\alpha_k) Q_k(x,a) + \alpha_k \BellmanOp Q_k(x,a) ,
\label{eq:RL}
\end{equation}
where $\alpha_k$ is the learning rate for iteration $k$.
At each timestep, \eqref{eq:RL} is iteratively applied to
the $Q$-function at the current state and action.

We note that each of the algorithms from the OpenAI Gym implements a form of the $\epsilon$-greedy method (e.g., occasionally picking a random action or using a randomly perturbed $Q$-function for determining the action) to enable some form of exploration in addition to the exploitation-based search of the
optimal policy using the $Q$-function.
Our experiments were therefore repeated over a wide range of values for $\epsilon$, where we found that the relative performance trends of the different approaches did not depend significantly on the amount of exploration under the $\epsilon$-greedy algorithm.
In particular, the same performance trends were observed over a wide range of $\epsilon$ values and hence we present results based on the default value of $\epsilon$ used in the reference codebase.

In short, as an overall general summary of the experimental results for each of the Lunar Lander, Mountain Car and Cart Pole problems considered, our control-based RL approach (CBRL) provides significantly better performance over that exhibited under the classical $Q$-learning approach with the Bellman operator (QLBO) in both training and testing phases.

\subsection{Lunar Lander}
\label{app:lunarlander}
This problem is discussed in
Brockman et al.~\cite{AIgym}.
In each state, characterized by an $8$-dimensional state
vector, there are $4$ possible discrete actions (left, right, vertical, no thrusters).
The goal is to maximize the cumulative reward (score) comprising positive points for successful degrees of landing and negative points for fuel usage and crashing.
For the
QLBO experiments,
the $6$ continuous state variables are each discretized into $4$ bins.
For the CBRL experiments, we consider $4$ different simple control models:
\begin{enumerate}
    \item Linear;
    \item Piecewise-Linear (PWL) Symmetric with $M=2$, $B_1=B_2$ except for $[B_1]_{*,1}=-[B_2]_{*,1}$, where $[B]_{*,1}$ denotes the first column of matrix $B$, and $b_1=b_2$;
    \item Piecewise-Linear (PWL) Symmetric with $M=4$, $B_1=B_4$ except for $[B_1]_{*,1}=-[B_4]_{*,1}$, $b_1=b_4$, $B_2=B_3$ except for $[B_2]_{*,1}=-[B_3]_{*,1}$, and $b_2=b_3$;
    \item Nonlinear with $g(x)=cx_1^2$ for scalar $x_1^2$.
\end{enumerate}

\begin{figure}[htbp]
\centering
{\includegraphics[width=0.9\textwidth]{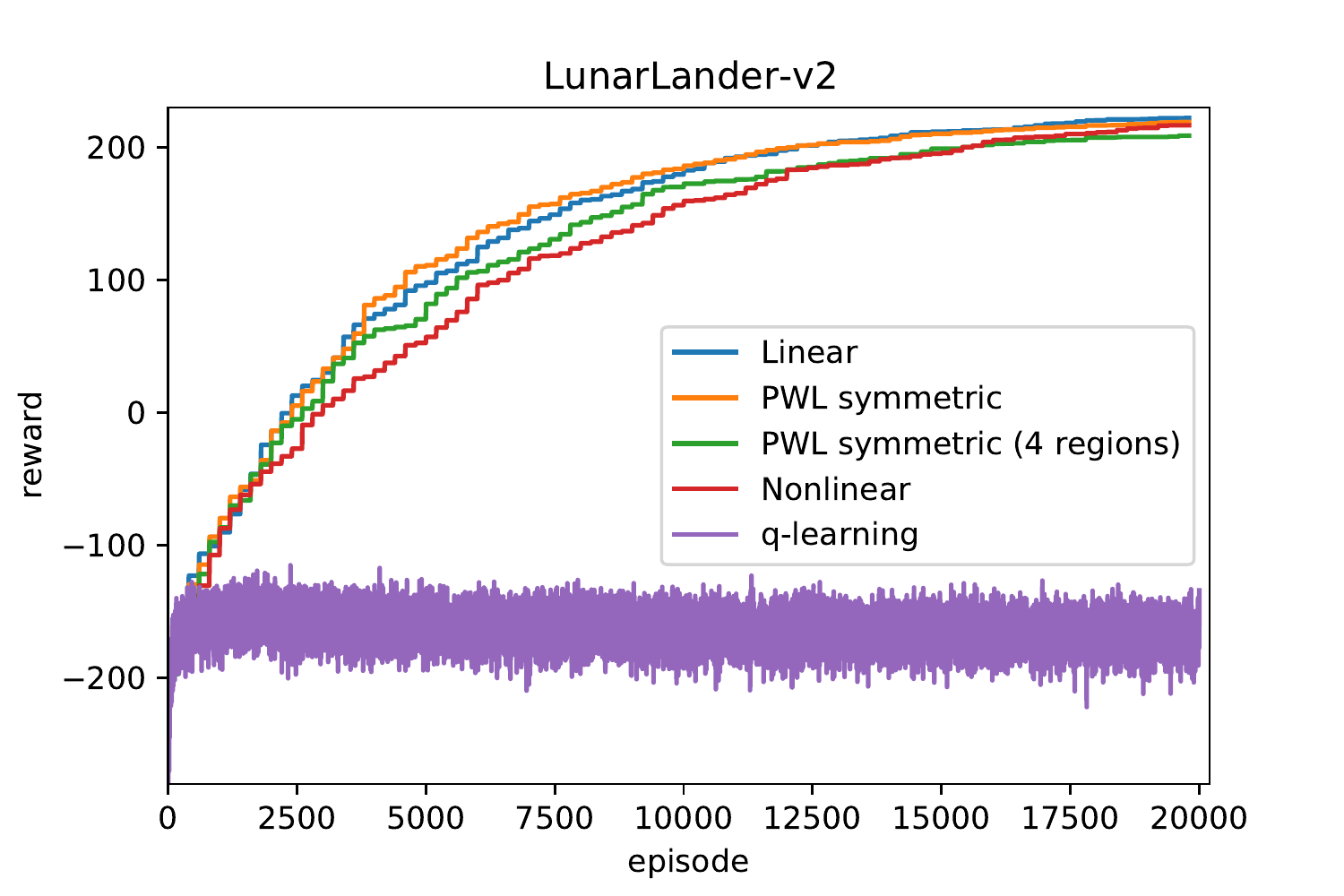}}
\caption{Performance Comparison of Various Instances of Our Control-Based RL Approach and Classical $Q$-Learning with the Bellman Operator, Average Lunar Lander Score over $50$ Trials, During Training Phase.}
\label{fig:app:lunarlander}
\end{figure}

Figure~\ref{fig:app:lunarlander} plots the score results, averaged over $50$ experimental trials, as a function of the number of episodes for CBRL under the $4$ simple control models and for QLBO during the training phase.
We observe that QLBO consistently exhibits significantly negative average scores with a trend over many episodes that remains flat.
In strong contrast, our CBRL with each of the simple control models finds control policies that achieve average scores above $+200$ within $20,000$ episodes with trends continuing to slightly improve.
We further observe that all $4$ CBRL result curves are quite similar.

Table \ref{tbl:lunarlander} presents
the average score over $1000$ episodes across the $50$ trials for CBRL under the $4$ simple control models and for QLBO during the testing phase, together with the corresponding 95\% confidence intervals.
We again observe that the best average scores are by far obtained under the simple CBRL control models, with Linear considerably above $+200$ and QLBO below $-200$,
and that the confidence intervals for all cases are quite small.
We further observe the performance orderings among the simple CBRL control models with Linear providing the best results, followed by Nonlinear, PWL Symmetric ($M=4$), and PWL Symmetric ($M=2$).
Note that the Linear and PWL Symmetric ($M=2$) controllers have the same number of parameters and they also provide the highest rewards during training; while the other CBRL controllers eventually approach these levels of reward performance during training, they do so more slowly due to the increased degrees of freedom (parameters). For the testing results, the CBRL controllers that provide better reward performance also have smaller confidence intervals, suggesting that they are closer to an optimal controller.
Furthermore, the Linear controller provides  the best performance, suggesting that the Nonlinear and PWL controllers require further searching of the decision space in the training phase to identify the corresponding optimal controller for the testing phase.
In particular, one can imagine that PWL controllers (and piecewise Nonlinear controllers) under appropriate partitionings should provide superior performance over that of the Linear controller for problems such as Lunar Lander, but this requires better searching of the decision space in the training phase to identify the corresponding optimal PWL controller for the testing phase, possibly along the lines of Section~\ref{sec:control+MDP}.

\begin{table}[htbp]
    \centering
    \begin{tabular}{|c|c|c|c|c|c|}
    \hline
         \shortstack{Control\\Model} & Linear & PWL Sym ($M=2$) & PWL Sym ($M=4$) & Nonlinear & Q-learning \\
         \hline\hline
        \shortstack{Score\\ $ $ } & {\bf \shortstack{ 216.05  \\ $\pm$  0.27\%}} & \shortstack{$116.26$\\$\pm 1.43\%$}  & \shortstack{$135.41$ \\ $\pm 0.94\% $}&\shortstack{$168.16$\\ $\pm 0.56\% $}&\shortstack{ $-203.21 $ \\ $\pm 0.51\%$} \\ \hline
    \end{tabular}
    \caption{Average Lunar Lander Scores Over $1000$ Episodes Across $50$ Trials for Instances of Our Control-Based RL Approach and Classical $Q$-Learning with the Bellman Operator, including $95$\% Confidence Intervals, During Testing Phase.}
    \label{tbl:lunarlander}
\end{table}

\subsection{Mountain Car}
\label{app:mountaincar}
This problem is first discussed in \cite{mountaincar}.
In each state, characterized by a $2$-dimensional state vector, there are $3$ possible actions (forward, backward, neutral).
The goal is to maximize the score representing the negation of the number of time steps needed to solve the problem (i.e., minimize number of steps to solve) over episodes of up to $200$ steps.
For the QLBO experiments, the state space is discretized into a $40\times 40$ grid.
For the CBRL experiments, we consider $2$ simple control models:
\begin{enumerate}
    \item Linear;
    \item Nonlinear with $g(x)$ as a second-order multivariate polynomial on the state variables.
\end{enumerate}

\begin{figure}[htbp]
\centering
{\includegraphics[width=0.9\textwidth]{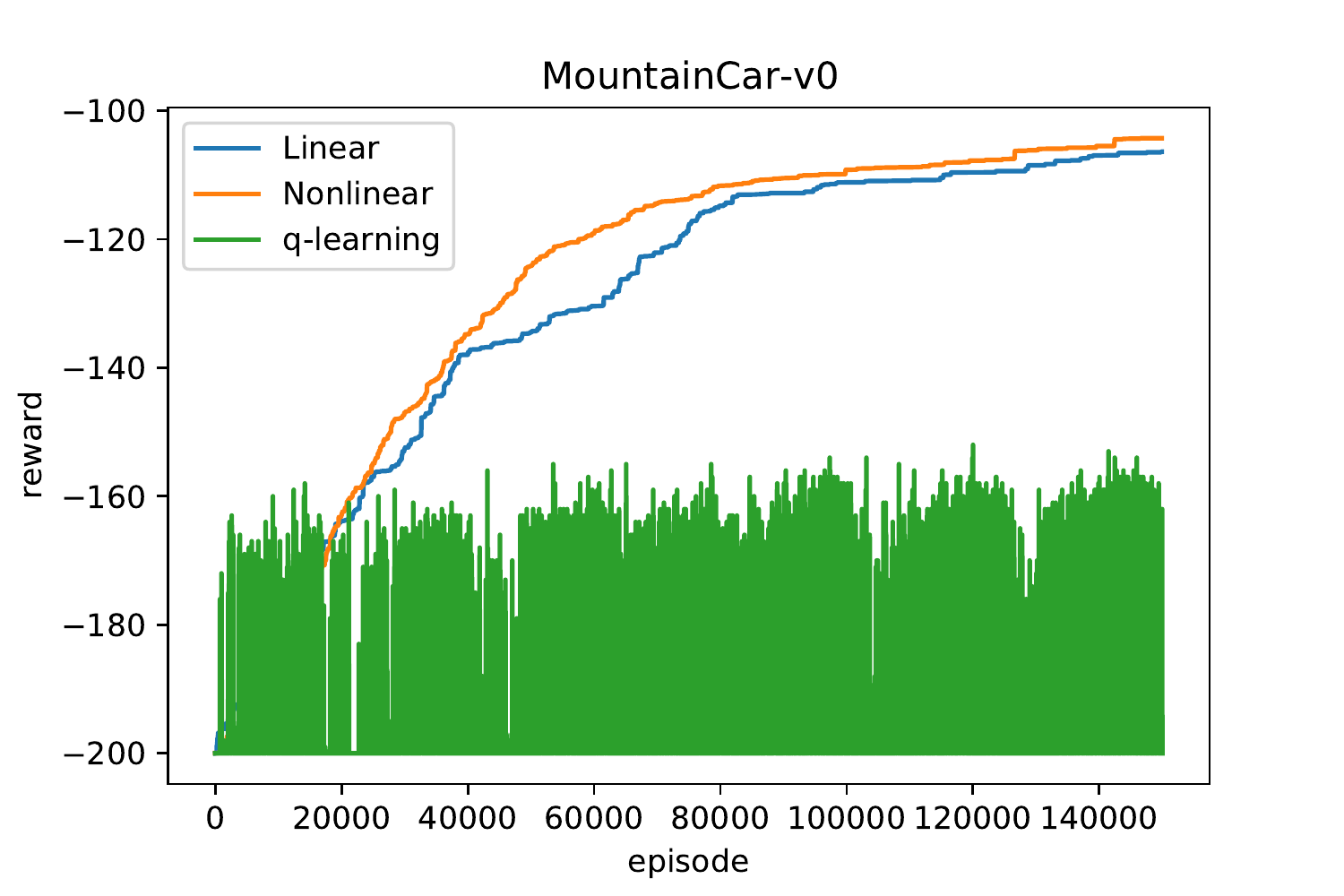}}
\caption{Performance Comparison of Various Instances of Our Control-Based RL Approach and Classical $Q$-Learning with the Bellman Operator, Average Mountain Car Score over $50$ Trials, During Training Phase.}
\label{fig:app:mountaincar}
\end{figure}

Figure~\ref{fig:app:mountaincar} plots the score results, averaged over $50$ experimental trials, as a function of the number of episodes for CBRL under the $2$ simple control models and for QLBO during the training phase.
We observe that the average scores for QLBO continually hover 
within the range of $[-160,-200]$,
implying that this algorithm is
rarely successful in solving the problem over a very large number of episodes across the $50$ trials.
In strong contrast, the average scores for CBRL with the simple Linear and Nonlinear control models both follow a similar path towards $-110$, which is the point at which the problem is considered to be solved.
We further observe the noticeable performance improvements under the Nonlinear controller over the Linear controller during the training phase.

Table \ref{tbl:mountaincar} presents the average score over $1000$ episodes across the $50$ trials for CBRL under the $2$ simple control models and for QLBO during the testing phase, together with the corresponding 95\% confidence intervals.
We again observe that the best average scores are by far obtained under the simple CBRL control models, with Linear and Nonlinear both solving the problem and QLBO still far away,
and that the confidence intervals for all cases are quite small.
We further observe the performance orderings among the simple CBRL control models with Nonlinear slightly outperforming Linear, but the differences are very small.
While the Nonlinear controller outperforms the Linear controller during both the training and testing phases, both simple CBRL control models perform extremely well, they completely solve the problem, and they significantly outperform QLBO.

\begin{table}[htbp]
    \centering
    \begin{tabular}{|c|c|c|c|}
    \hline
         Control Model  & Linear & Nonlinear & Q-learning \\
         \hline\hline
         Score &  -105.72 $\pm$ 0.11\%  & {\bf -105.48 $\pm$ 0.11\%} & -129.68 $\pm$ 0.21\% \\\hline
    \end{tabular}
    \caption{Average Mountain Car Scores Over $1000$ Episodes Across $50$ Trials for Instances of Our Control-Based RL Approach and Classical $Q$-Learning with the Bellman Operator, including $95$\% Confidence Intervals, During Testing Phase.}
    \label{tbl:mountaincar}
\end{table}

\subsection{Cart Pole}
\label{app:cartpole}
This problem is discussed in Barto et al.~\cite{cartpole}.
In each state, characterized by a $4$-dimensional state vector, there are $2$ possible discrete actions (push left, push right).
The goal is to maximize the score representing the number of steps that the cart pole stays upright before either falling over or going out of bounds.
For the QLBO experiments, the position and velocity are discretized into $8$ bins whereas the angle and angular velocity are discretized into $10$ bins.
For the CBRL experiments, we consider $2$ different simple control models:
\begin{enumerate}
    \item Linear;
    \item Nonlinear with $g(x)$ as a second-order multivariate polynomial on the state variables.
\end{enumerate}
With a score of $200$, the problem is considered solved and the simulation ends.

\begin{figure}[htbp]
\centering
{\includegraphics[width=0.9\textwidth]{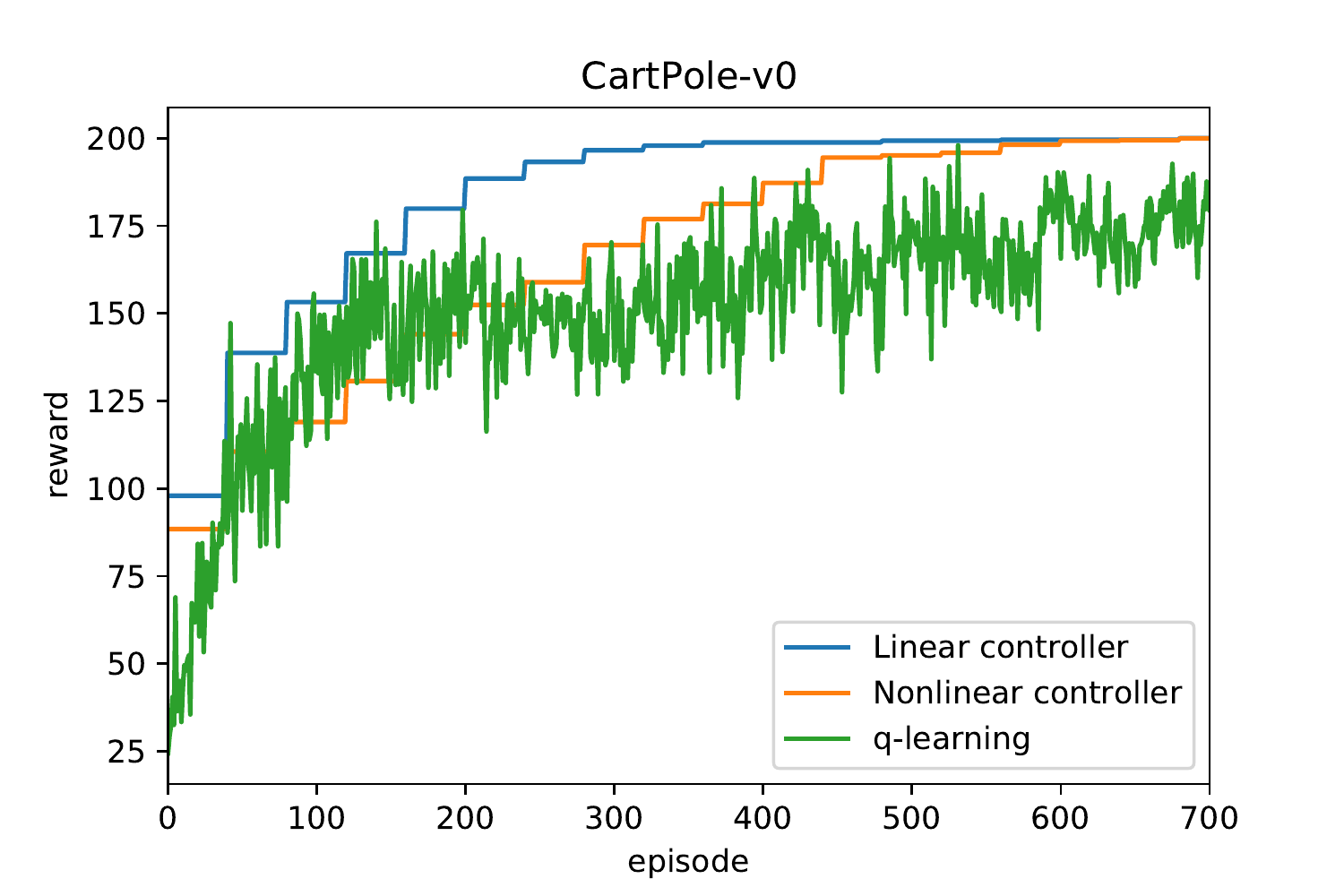}}
\caption{Performance Comparison of Various Instances of Our Control-Based RL Approach and Classical $Q$-Learning with the Bellman Operator, Average Cart Pole Score over $50$ Trials, During Training Phase.}
\label{fig:app:cartpole}
\end{figure}

Figure~\ref{fig:app:cartpole} plots the score results, averaged over $50$ experimental trials, as a function of the number of episodes for CBRL under the $2$ simple control models and for QLBO during the training phase.
We observe that CBRL, with both the simple Linear and Nonlinear control models, quickly finds an optimal control policy that solves the problem within a few hundred episodes, whereas QLBO continues to oscillate well below the maximal score of $200$ even for this relatively simple problem.

Table \ref{tbl:cartpole} presents
the average score over $1000$ episodes across the $50$ trials for CBRL under the $2$ simple control models and for QLBO during the testing phase, together with the corresponding 95\% confidence intervals.
We again observe that the best average scores are by far obtained under the simple CBRL control models, with Linear and Nonlinear both solving the problem while having confidence intervals of essentially zero and with QLBO providing poorer performance and still a considerable distance away from optimal while having a very small confidence interval. In addition, the Linear controller with less degrees of freedom reaches higher scores more quickly than the Nonlinear controller during the training phase, but both controllers reach a score of $200$ after a few hundred episodes.

\begin{table}[htbp]
    \centering
    \begin{tabular}{|c|c|c|c|}
    \hline
         Controller/policy  & Linear & Nonlinear & Q-learning \\
         \hline\hline
         Reward & {\bf 200.0 $\pm$ 0\%} & {\bf 200.0 $\pm$ 0\%} & 189.2 $\pm$ 0.24\%\\\hline
    \end{tabular}
    \caption{Average Cart Pole Scores Over $1000$ Episodes Across $50$ Trials for Instances of Our Control-Based RL Approach and Classical $Q$-Learning with the Bellman Operator, including $95$\% Confidence Intervals, During Testing Phase.}
    \label{tbl:cartpole}
\end{table}

\end{document}